%% file: main.tex
\documentclass[10pt,twocolumn,letterpaper]{article}

\usepackage{cvpr}              
\usepackage{xspace}
\usepackage{adjustbox}
\usepackage{booktabs}
\usepackage{graphicx}
\usepackage{caption}
\usepackage{subcaption}
\usepackage{color}

\input{preamble}

\definecolor{cvprblue}{rgb}{0.21,0.49,0.74}
\usepackage[pagebackref,breaklinks,colorlinks,citecolor=cvprblue]{hyperref}

\def\paperID{*****} 
\def\confName{CVPR}
\def\confYear{2024}

\title{ViT-MUL: A Baseline Study on Recent Machine Unlearning Methods Applied to Vision Transformers}

\author{
    \textbf{Ikhyun Cho} \\ 
    University of Illinois at\\
    Urbana-Champaign  \\
    {\tt\small ihcho2@illinois.edu}
    \and
    \textbf{Changyeon Park} \\
    Seoul National University \\
    {\tt\small blackco@snu.ac.kr}
    \and 
    \textbf{Julia Hockenmaier} \\
    University of Illinois at \\
    Urbana-Champaign \\
    {\tt\small juliahmr@illinois.edu}
}

\begin{document}
\maketitle
\input{000Abstract} 

\section{Introduction}
\input{010Intro}
\label{010intro}

\section{Related Work}
\input{020Related}
\label{020related}

\section{Experiments}
\input{040Experiments}
\label{040experiments}

\section{Conclusion}
\input{050Conclusion}
\label{050conclusion}

{
    \small
    \bibliographystyle{ieeenat_fullname}
    \bibliography{main}
}


\end{document}

%% file: preamble.tex
\usepackage[dvipsnames]{xcolor}


%% file: 000Abstract.tex
\begin{abstract}
Machine unlearning (MUL) is an arising field in machine learning that seeks to erase the learned information of specific training data points from a trained model.
Despite the recent active research in MUL within computer vision, the majority of work has focused on ResNet-based models. Given that Vision Transformers (ViT) have become the predominant model architecture, a detailed study of MUL specifically tailored to ViT is essential. In this paper, we present comprehensive experiments on ViTs using recent MUL algorithms and datasets. We anticipate that our experiments, ablation studies, and findings could provide valuable insights and inspire further research in this field.
\end{abstract}

%% file: 010Intro.tex
Machine unlearning (MUL), aiming to eliminate knowledge of specific training data points stored in a pre-trained model, has emerged as a significant research area in deep learning. This trend is driven by the growing privacy concerns associated with large pre-trained models \cite{choi2023towards}. Furthermore, laws such as \textit{The Right to be Forgotten} \cite{voigt2017eu} and the \textit{California Consumer Privacy Act} \cite{cali} grant users the authority to request companies to delete their privacy-related information from pre-trained models. 

A straightforward solution to Machine Unlearning (MUL) is to re-train the model from scratch using the modified training data, where the data points that need to be forgotten is excluded. However, this approach is often computationally too expensive, given the large volume of training data. Therefore, the search for an efficient way to \textit{unlearn} the pre-trained model, as opposed to retraining from scratch, has been a subject of extensive recent research.

Despite recent active research in MUL, the majority of studies have focused on ResNet-based models \cite{kurmanji2023towards, choi2023towards, jung2024attack, bourtoule2021machine}. However, Vision Transformer (ViT) has emerged as the dominant model architecture in various areas of computer vision. Hence, there is a crucial need for MUL studies that specifically target ViT models. In response to this need, we conduct comprehensive machine unlearning experiments on ViT models using the recently proposed MUL algorithms and datasets \cite{choi2023towards}. Specifically, we utilize two most widely-used ViT models, ViT-base and ViT-large, applying and analyzing recent machine unlearning algorithms on these architectures. We anticipate that our experiments, ablation studies, and findings could offer valuable insights and motivate further research in this field. Code is available at \url{https://github.com/ihcho2/ViTMUL}.


%% file: 020Related.tex
\subsection{Vision Transformer (ViT)}
Transformers \cite{vaswani2017attention}, initially prevalent in natural language processing \cite{devlin2018bert, liu2019roberta, sun2019patient, cho2023sir, cho2022pea, piao2022sensimix}, have now emerged as a dominant model architecture in computer vision as well. 
Specifically, ViT \cite{dosovitskiy2010image} forms the basis of current state-of-the-art techniques in various tasks (e.g., image classification, image segmentation, image captioning e.t.c.) \cite{radford2021learning, mokady2021clipcap, yu2023noisynn}. 
Unlike traditional approaches, ViT exclusively employs a transformer architecture. This process involves dividing images into fixed-size patches (tokens), adding learnable position embeddings to each token, which are then processed through multiple layers of self-attention-based transformers. This method enables the learning of relationships between the segmented image parts in a parallelizable way, ultimately leading to an effective image understanding. ViT models, pre-trained on millions of image-context pairs, have now become the de facto models used in a variety of fields in computer vision \cite{radford2021learning, mokady2021clipcap}. Therefore, it is imperative to test machine unlearning methods on ViT-based models.
 
\subsection{Machine Unlearning}

\noindent\textbf{Background}\quad The concept of Machine Unlearning was initially introduced by \cite{cao2015towards}, who utilized statistical query learning in its development. Subsequently, a significant body of research, as seen in studies \cite{golatkar2020eternal, goel2022towards}, has concentrated on devising machine unlearning methodologies specifically for deep neural networks. In this paper, we undertake a comparative analysis of various machine unlearning approaches, with the aim of establishing a foundational baseline for machine unlearning in ViT systems.

\noindent\textbf{Formal definition and metrics}\quad Given a pre-trained model $\theta_0$ initially trained on $D_{T}$, suppose we want to make $\theta_0$ forget a set of data $D_F$ that is part of the training data (i.e., $D_F \subset D_{T}$). Here, $D_F$ is referred to as the forget set, and the complement $D_R (=D_T \cap D_F^c)$ is termed the retain set. The goal of machine unlearning is to create an unlearning algorithm $U$ in such a way that, upon application, the unlearned model, $\theta^* = U(\theta_0, D_F, D_R)$, effectively forgets $D_F$. Machine unlearning algorithms are typically assessed using two key metrics: \textit{utility} and \textit{forgetting performance}.

Utility is evaluated by assessing the model's accuracy on a distinct test set, $D_{Test}$. A valuable unlearning algorithm should ensure that the overall performance of the model does not degrade significantly. Hence, we expect $\theta^*$ to have a similar (or ideally an improved) test accuracy compared to $\theta_0$.

To assess the forgetting performance, an additional distinct test set $D_{Unseen}$ is used. At a high level, the objective is to ensure that the unlearned model's behavior on the forget set $D_F$ closely resembles its behavior on $D_{Unseen}$. To achieve this, a classifier, such as a regression model, is typically trained to distinguish between the outputs, $\theta^*(D_F)$ and $\theta^*(D_{Unseen}$). That is, we collect the losses from $\theta^*(D_F)$ and $\theta^*(D_{Unseen})$, and then train a regression classifier to distinguish between them. If the classifier's accuracy is close to 50\% (i.e., the classifier is unable to distinguish them), it indicates a high similarity between the outputs $\theta^*(D_F)$ and $\theta^*(D_{Unseen}$), suggesting that the model has effective forgot $D_F$ (more details can be found in \cite{choi2023towards}).

Intuitively, if a model efficiently unlearns specific data, the general performance is likely to decrease, especially when dealing with a large amount of forget data. Consistent with this intuition, a general trade-off exists between the model's utility and its forgetting performance \cite{jung2024attack, choi2023towards}. The primary goal of MUL is to derive a decent trade-off between them.

\subsection{Instance-based Machine Unlearning}
\label{023instance}
Most previous MUL studies have used conventional computer vision datasets such as MNIST \cite{mnist}, CIFAR-10 \cite{CIFAR}, and SVHN \cite{SVHN}. Prior research focused on unlearning specific class(es) (e.g., unlearning numbers of 9 in MNIST). However, recent observations indicate that this setting does not align well with real-world applications \cite{choi2023towards}. In practice, the need often arises to forget specific instances (individuals) that may have different labels, rather than an entire class. Acknowledging this, \citeauthor{choi2023towards} \cite{choi2023towards} introduced two new machine unlearning benchmark datasets: MUFAC and MUCAC. In this paper, our focus is on instance-based machine unlearning, as it is more aligned with real-world scenarios.

\noindent\textbf{MUFAC (Machine Unlearning for Facial Age Classifier)}\quad \citeauthor{choi2023towards} \cite{choi2023towards} introduced a multi-class age classification dataset, MUFAC, comprising Asian facial images with annotated labels. The labels categorize individuals into eight age groups. Unlike previous machine unlearning tasks that aim to forget specific class(es), MUFAC focuses on unlearning a group of individuals with various class labels. Similar to previous MUL tasks, the objective is to achieve a balanced trade-off between test accuracy and forgetting quality.

\noindent\textbf{MUCAC (Machine Unlearning for Celebrity Attribute Classifier)}\quad 
\citeauthor{choi2023towards} \cite{choi2023towards} introduced another MUL dataset derived from CelebA \cite{liu2018large}, called MUCAC, which involves a multi-label facial classification task. The labels contain three attributes: gender (male/female), age (old/young), and expression (smiling/unsmiling). Statistical details of MUFAC and MUCAC can be found in Section~\ref{031datasets}.

\subsection{Baselines}
We provide an overview of well-known machine unlearning algorithms widely used in recent days below.

\noindent\textbf{Fine-Tuning}\quad Refining the original model through fine-tuning, utilizing solely the retain set, can be an effective approach. During this learning process, employing a marginally higher learning rate can enhance generalization. This adjustment may lead to more effective forgetting of the data intended to be unlearned \cite{golatkar2020eternal}.

\noindent\textbf{Catastrophically Forgetting-k}\quad The concept of Catastrophically Forgetting in the last k layers (CF-k), as proposed by \cite{goel2022towards}, involves fine-tuning only the last k layers while keeping the rest unchanged. This method leverages the phenomenon of catastrophic forgetting in machine learning models, as documented in \cite{french1999catastrophic}. Catastrophic forgetting occurs when a model is repeatedly updated without retraining on previously learned data, leading to a gradual loss of information related to that data. By focusing the learning process on the last k layers, this approach requires fewer training epochs and is able to maintain the utility of the model more effectively.

\noindent\textbf{Advanced Negative Gradient}\quad Introduced by \cite{choi2023towards}, Advanced Negative Gradient (AdvNegGrad) is an enhanced version of Negative Gradient (NegGrad) \cite{golatkar2020eternal}. While NegGrad applies gradient ascent using the forget set to increase loss, leading to data oblivion, AdvNegGrad integrates the joint loss of fine-tuning with NegGrad's approach within the same training batches. 

\noindent\textbf{Unlearning by Selective Impair and Repair}\quad The Unlearning by Selective Impair and Repair (UNSIR) method, proposed by \cite{tarun2023fast}, initially designed to forget specific classes in a model, introduces disruptive noise to negatively impact the model's weights during the learning phase. After this corruption, the model undergoes fine-tuning with a retain set to rectify these changes. However, unlike its original purpose, in this paper \cite{choi2023towards}, UNSIR is adapted to forget specific individual data points. This is achieved by learning a synthesized noise that maximizes the difference from the data to be forgotten, followed by fine-tuning to correct the altered weights.

\noindent\textbf{SCalable Remembering and Unlearning unBound} In this context, a technique is used where the knowledge of the forget set is removed by employing a stochastic initialization model, which serves as a student model. This method differs from joint training as it uses a ``bad teacher'' concept to eliminate the impact of the forget set. However, this approach of maximizing the distance between the student and teacher models for the forget set can adversely affect the performance on the retain set. To address this issue, \cite{kurmanji2023towards} introduced SCalable Remembering and Unlearning unBound (SCRUB). This method aims to maintain the student model's closeness to the teacher on the retain set while distancing it on the forget set. 

\noindent\textbf{Attack-and-Reset}\quad Presented by \cite{jung2024attack}, Attack-and-Reset (ARU) method identifies and re-initializes parameters in a model that are prone to overfitting on the forget set. It determines these parameters by measuring the gradient differences between the original forget images and noise. If this gradient discrepancy is small, it indicates that these parameters are responsible to the model's overfitting. By re-initializing these specific parameters and applying fine-tuning afterwards, ARU leads to the model forgetting the data in the forget set.

In this paper, we apply the aforementioned baseline algorithms to Vision Transformer (ViT)-based models and assess their performance in an instance-based machine unlearning setting using the MUFAC and MUCAC datasets.

%% file: 040Experiments.tex
\subsection{Datasets}
\label{031datasets}
As explained in Section~\ref{023instance}, we focus on unlearning instances rather than unlearning a class(es), as it better aligns with real-world applications. Hence, we use the recently introduced datasets, MUFAC and MUCAC, provided by \citeauthor{choi2023towards} \cite{choi2023towards}. Statistics of these datasets are summarized in Table~\ref{tab:STATS}. All images in MUFAC and MUCAC have a resolution of 128×128, focusing on the facial region. 

\begin{table}[ht!]
\centering
\begin{tabular}{lcc}
\toprule
 &  MUFAC   &  MUCAC  \\
\midrule
Train dataset & 10,025& 25,846\\
Test dataset & 1,539  &2,053\\
Forget dataset & 1,500 & 10,135 \\
Retain dataset & 8,525 & 15,711\\
Unseen dataset & 1,504 &2,001 \\
\bottomrule
\end{tabular}
\caption{Overall statistics of MUFAC and MUCAC.}
\label{tab:STATS}
\end{table}

\subsection{Experimental Settings}
We use two of the most widely used ViT models, ViT-Base and ViT-Large. For faster convergence and better overall performance, we use the pre-trained models (ViT-B-16 and ViT-L-14) provided by \citeauthor{radford2021learning} \cite{radford2021learning}. We follow the default hyper-parameter settings from the official repository \cite{radford2021learning} (i.e., optimizer of adamW, learning rate of 1e-5, batch size of 64, patch size of 16x16 for ViT-B-16 and 14x14 for ViT-L-14). We use 30 epochs for unlearning, recording the best outcome based on the NoMUS score. For accurate evaluations, we use 5 random seeds equally for all algorithms and report the average along with standard deviations. 

\subsection{Overall Results}
\begin{table*}[h!]
  \centering
  \begin{tabular}{l|ccc|ccc}
    \toprule
    &  & MUFAC & & & MUCAC & \\
    \midrule
    Model &  Utility (\%, $\uparrow$) & Forget (\%, $\downarrow$) & NoMUS (\%, $\uparrow$) & Utility (\%, $\uparrow$) & Forget (\%, $\downarrow$) & NoMUS (\%, $\uparrow$) \\
    \midrule
    1. ResNet18 \\
    \midrule
    Pre-trained & 59.52 & 21.36 & 58.40 & 88.52 & 4.19 & 90.07\\
    \midrule
    Unlearning & & & & \\
    \, $\bullet$ Re-train & 47.34 ($\pm$0.84) & 3.09 ($\pm$1.06) & 70.58 ($\pm$0.92) & 87.62 ($\pm$3.38) & 3.03 ($\pm$1.52) & 90.77 ($\pm$1.68)\\
    \, $\bullet$ Finetune & 59.57 ($\pm$0.43) & 19.89 ($\pm$0.16) & 59.90 ($\pm$0.31) & 91.05 ($\pm$0.78) & 3.17 ($\pm$0.61) & 92.35 ($\pm$0.66)\\
    \, $\bullet$ CF-k & 59.42 ($\pm$0.55) & 20.11 ($\pm$0.28) & 59.60 ($\pm$0.39) & 91.96 ($\pm$0.14) & 4.29 ($\pm$0.55) & 91.69 ($\pm$0.58) \\
    \, $\bullet$ AdvNegGrad & 49.37 ($\pm$0.63) & 0.56 ($\pm$0.68) & 74.13 ($\pm$0.78) & 88.51 ($\pm$2.58) & 3.32 ($\pm$0.61) & 90.93 ($\pm$1.24) \\
    \, $\bullet$ UNSIR & 59.07 ($\pm$0.40) & 20.27 ($\pm$0.27) & 59.27 ($\pm$0.36) & 91.98 ($\pm$0.40) & 3.61 ($\pm$0.45) & 92.38 ($\pm$0.40)\\
    , $\bullet$ SCRUB & 52.45 ($\pm$1.05) & 0.99 ($\pm$0.73) & 75.23 ($\pm$0.71) & 90.09 ($\pm$1.61) & 2.62 ($\pm$1.11) & 92.43 ($\pm$0.49) \\
    , $\bullet$ ARU & 59.25 ($\pm$1.31) & 0.61 ($\pm$0.42) & 79.01 ($\pm$0.49) & 90.33 ($\pm$0.74) & 2.00 ($\pm$0.62)& 93.17 ($\pm$0.59) \\
    \midrule
    2. ViT-B-16 \\
    \midrule
    Pre-trained & 66.54 & 12.56 & 70.71 & 95.74 & 8.95 & 88.92 \\
    \midrule
    Unlearning & & & & \\
    \, $\bullet$ Re-train & 62.70 ($\pm$2.28) & 5.94 ($\pm$1.20) & 75.40 ($\pm$1.42) & 95.07 ($\pm$0.26) & 2.10 ($\pm$0.41) & 95.43 ($\pm$0.40)\\
    \, $\bullet$ Finetune & 64.78 ($\pm$0.79) & 2.03 ($\pm$1.05) & 80.36 ($\pm$0.81) & 94.50 ($\pm$0.71) & 3.30 ($\pm$0.39) & 93.95 ($\pm$0.18)\\
    \, $\bullet$ CF-k & 64.98 ($\pm$0.89) & 2.37 ($\pm$1.27) & 80.12 ($\pm$0.99) & 94.91 ($\pm$0.18) & 3.78 ($\pm$0.30) & 93.67 ($\pm$0.26) \\
    \, $\bullet$ AdvNegGrad & 63.10 ($\pm$1.26) & 1.04 ($\pm$0.97) & 80.51 ($\pm$0.85) & 94.08 ($\pm$0.13) & 0.42 ($\pm$0.43) & 96.62 ($\pm$0.44)\\
    \, $\bullet$ UNSIR & 64.93 ($\pm$0.90) & 2.26 ($\pm$0.76) & 80.20 ($\pm$0.74) & 94.72 ($\pm$0.31) & 3.73 ($\pm$0.58) & 93.62 ($\pm$0.58)\\
    , $\bullet$ SCRUB & 65.93 ($\pm$0.84) & 1.45 ($\pm$0.76) & 81.52 ($\pm$0.42) & 93.74 ($\pm$2.21) & 4.03 ($\pm$1.38) & 92.84 ($\pm$0.34) \\
    \, $\bullet$ ARU & 62.44 ($\pm$0.62) & 0.96 ($\pm$0.94) & 80.26 ($\pm$0.92) & 94.63 ($\pm$0.14) & 2.96 ($\pm$0.73) & 94.36 ($\pm$0.76)\\
    \midrule
    3. ViT-L-14 \\
    \midrule
    Pre-trained & 71.35 & 18.9 & 66.77 & 95.63 & 6.58 & 91.23 \\
    \midrule
    Unlearning & & & & \\
    \, $\bullet$ Re-train & 67.23 ($\pm$1.49) & 5.05 ($\pm$0.64) & 78.56 ($\pm$0.95) & 94.88 ($\pm$0.14) & 1.25 ($\pm$0.27) & 96.19 ($\pm$0.24) \\
    \, $\bullet$ Finetune & 67.75 ($\pm$1.40) & 8.34 ($\pm$1.14) & 75.53 ($\pm$0.56) & 94.97 ($\pm$0.21) & 4.22 ($\pm$0.27) & 93.27 ($\pm$0.24)\\
    \, $\bullet$ CF-k & 68.98 ($\pm$1.15) & 10.13 ($\pm$1.07) & 74.36 ($\pm$1.11) & 94.85 ($\pm$0.10) & 4.41 ($\pm$0.27) & 93.01 ($\pm$0.30)\\
    \, $\bullet$ AdvNegGrad & 66.33 ($\pm$1.84) & 2.76 ($\pm$1.24)  & 80.40 ($\pm$0.98) & 94.28 ($\pm$0.22) & 0.41 ($\pm$0.31) & 96.72 ($\pm$0.27)\\
    \, $\bullet$ UNSIR & 66.81 ($\pm$1.86) & 5.42 ($\pm$1.19) & 77.98 ($\pm$0.62) & 94.52 ($\pm$0.38) & 2.42 ($\pm$0.29) & 94.84 ($\pm$0.34)\\
    , $\bullet$ SCRUB & 67.28 ($\pm$3.38) & 2.92 ($\pm$2.06) & 80.72 ($\pm$1.14) & 94.42 ($\pm$0.89) & 3.51 ($\pm$0.99) & 93.70 ($\pm$0.60)\\
    \, $\bullet$ ARU & 65.07 ($\pm$1.38) & 0.53 ($\pm$0.46) & 82.01 ($\pm$0.59) & 94.98 ($\pm$0.14) & 2.87 ($\pm$0.39) & 94.62 ($\pm$0.37)\\
    \bottomrule
  \end{tabular}
  \caption{Overall results of unlearning experiments on MUFAC and MUCAC benchmarks using ViT-based models. $\uparrow/\downarrow$ indicates the metrics being larger/smaller the better, respectively.}
  \label{tab:overall}
\end{table*}

\begin{figure*}[h!]
\begin{subfigure}{.245\textwidth}
  \centering
  \includegraphics[width=\textwidth]{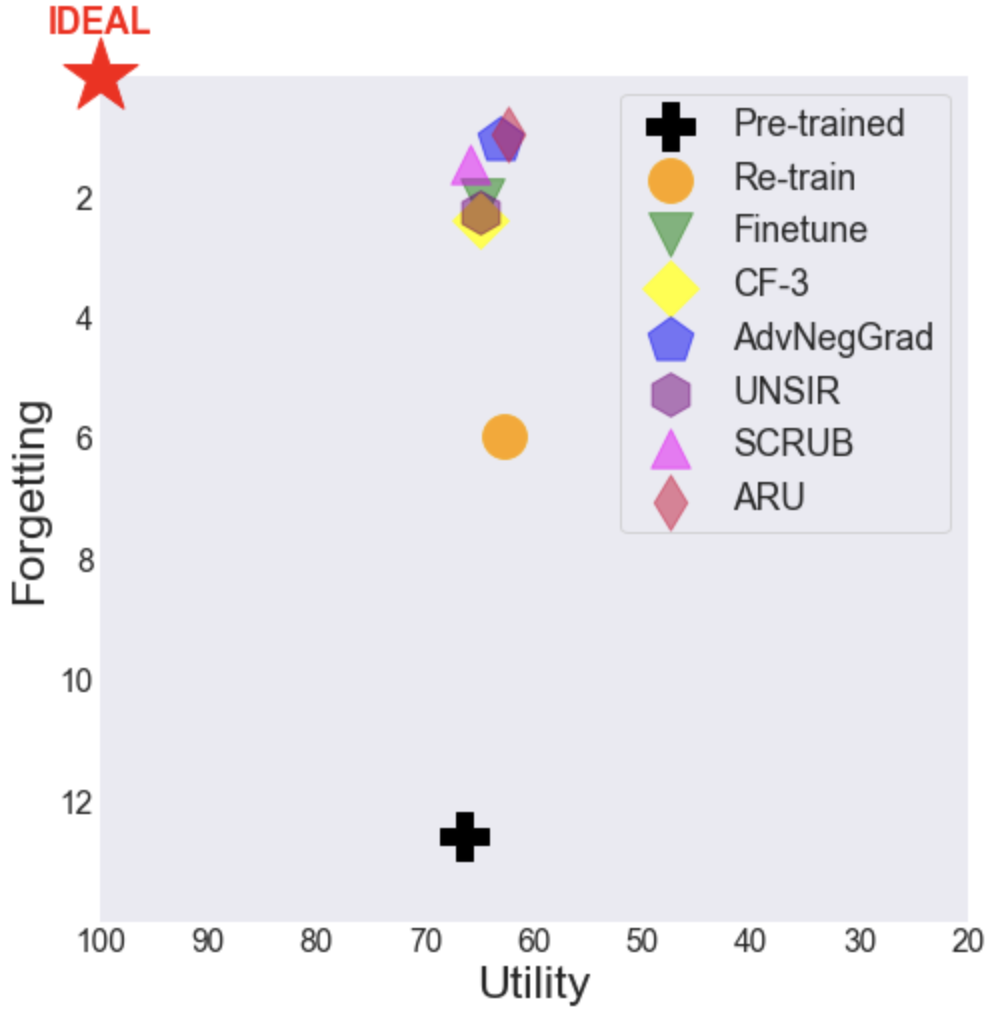}  
  \caption{Trade-off (ViT-B-16, MUFAC)}
  \label{fig:sub-first}
\end{subfigure}
\begin{subfigure}{.245\textwidth}
  \centering
  \includegraphics[width=\linewidth]{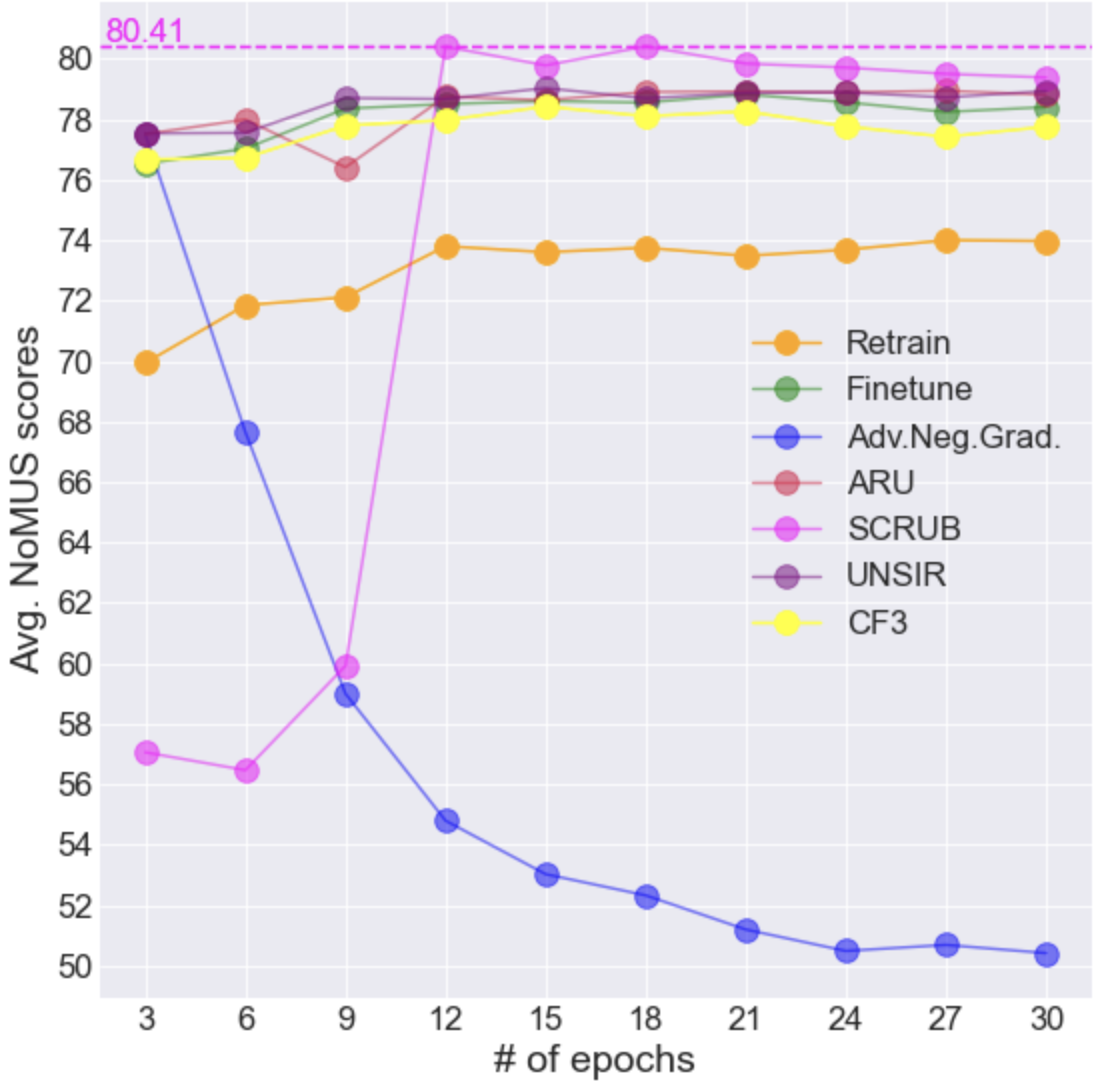}  
  \caption{NoMUS (ViT-B-16, MUFAC)}
  \label{fig:sub-second}
\end{subfigure}
\begin{subfigure}{.245\textwidth}
  \centering
  \includegraphics[width=\linewidth]{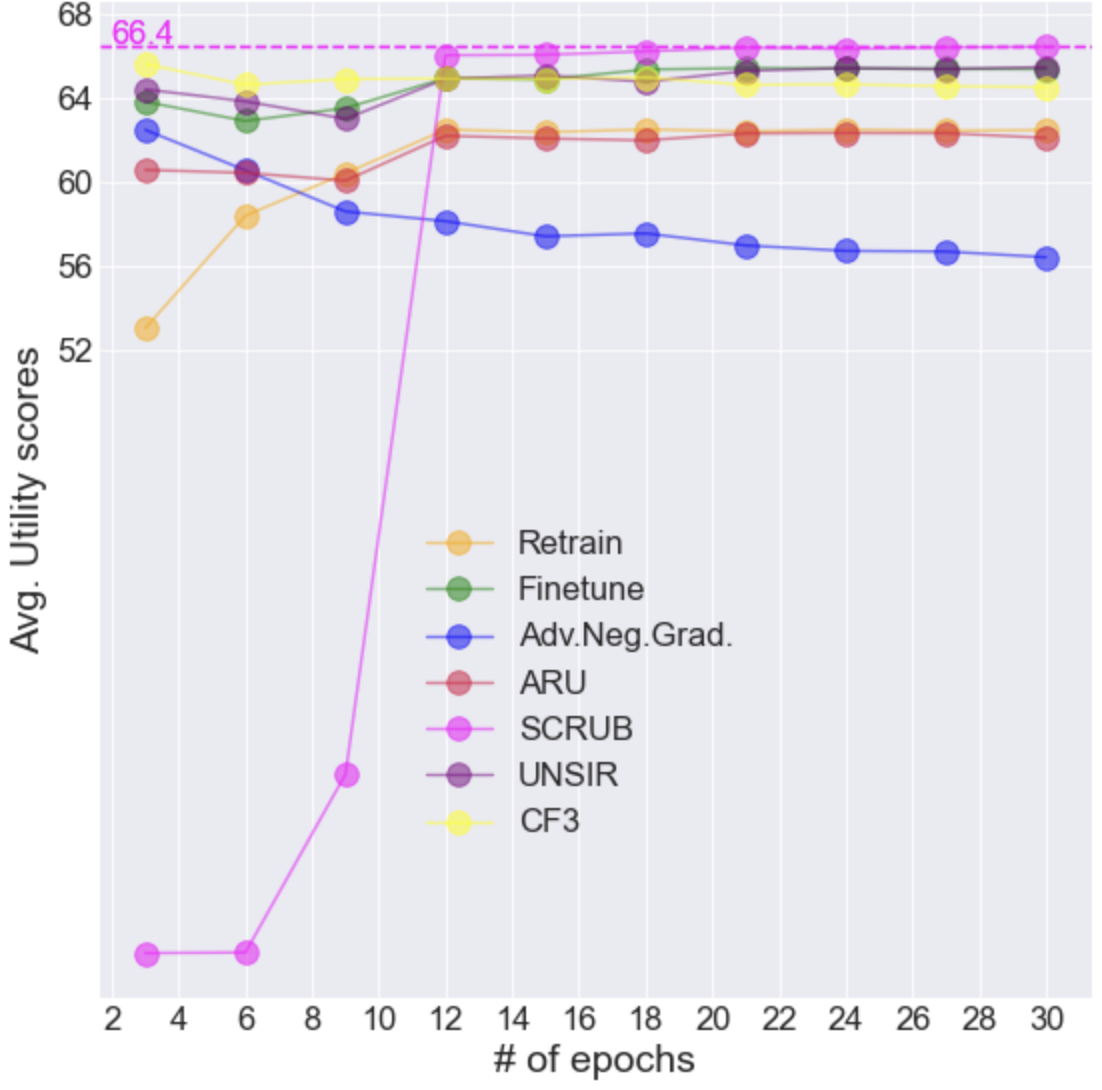}  
  \caption{Utility (ViT-B-16, MUFAC)}
  \label{fig:sub-second}
\end{subfigure}
\begin{subfigure}{.245\textwidth}
  \centering
  \includegraphics[width=\linewidth]{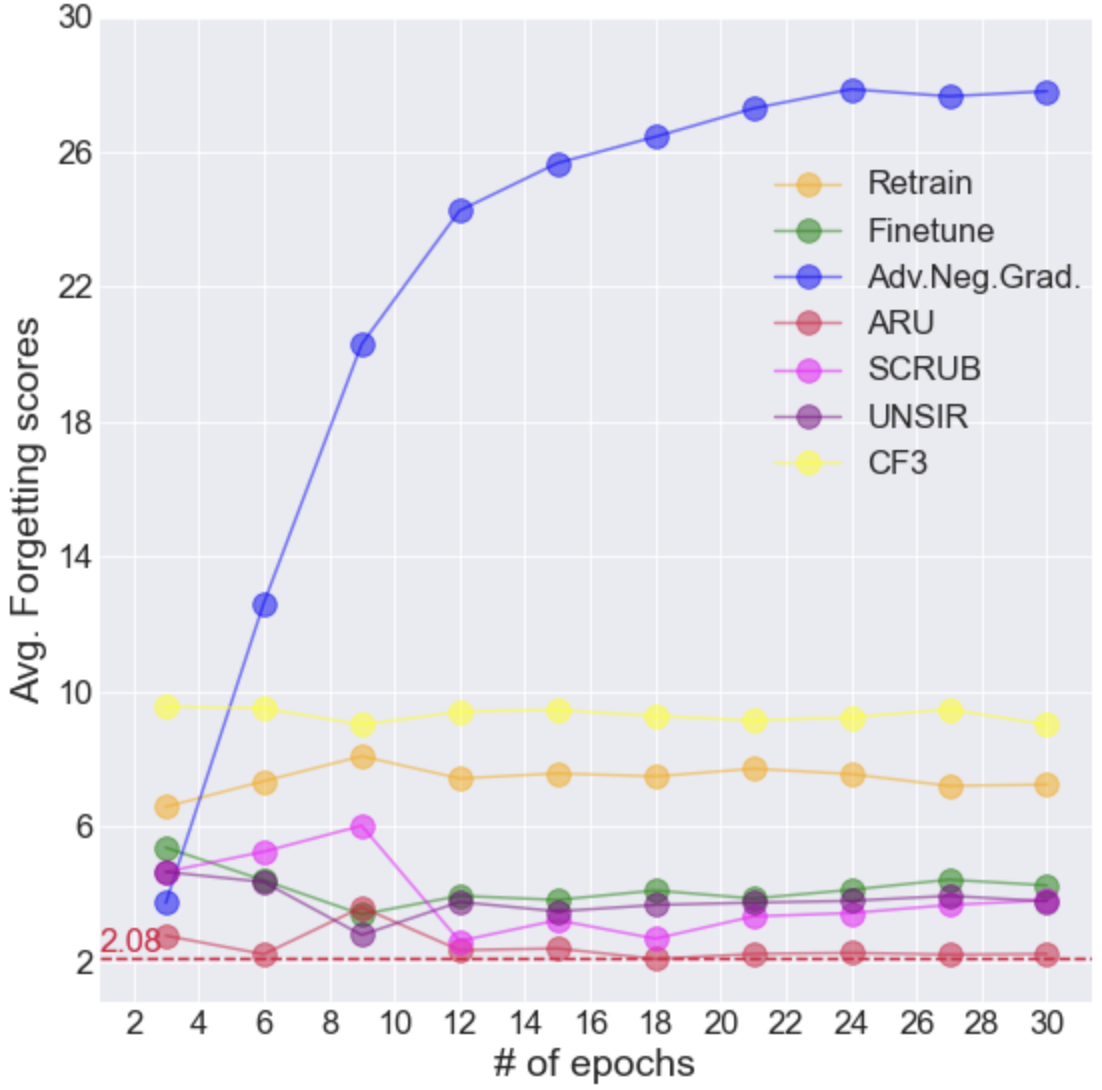}  
  \caption{Forgeting (ViT-B-16, MUfAC)}
  \label{fig:sub-second}
\end{subfigure}
\newline
\begin{subfigure}{.245\textwidth}
  \centering
  \includegraphics[width=\textwidth]{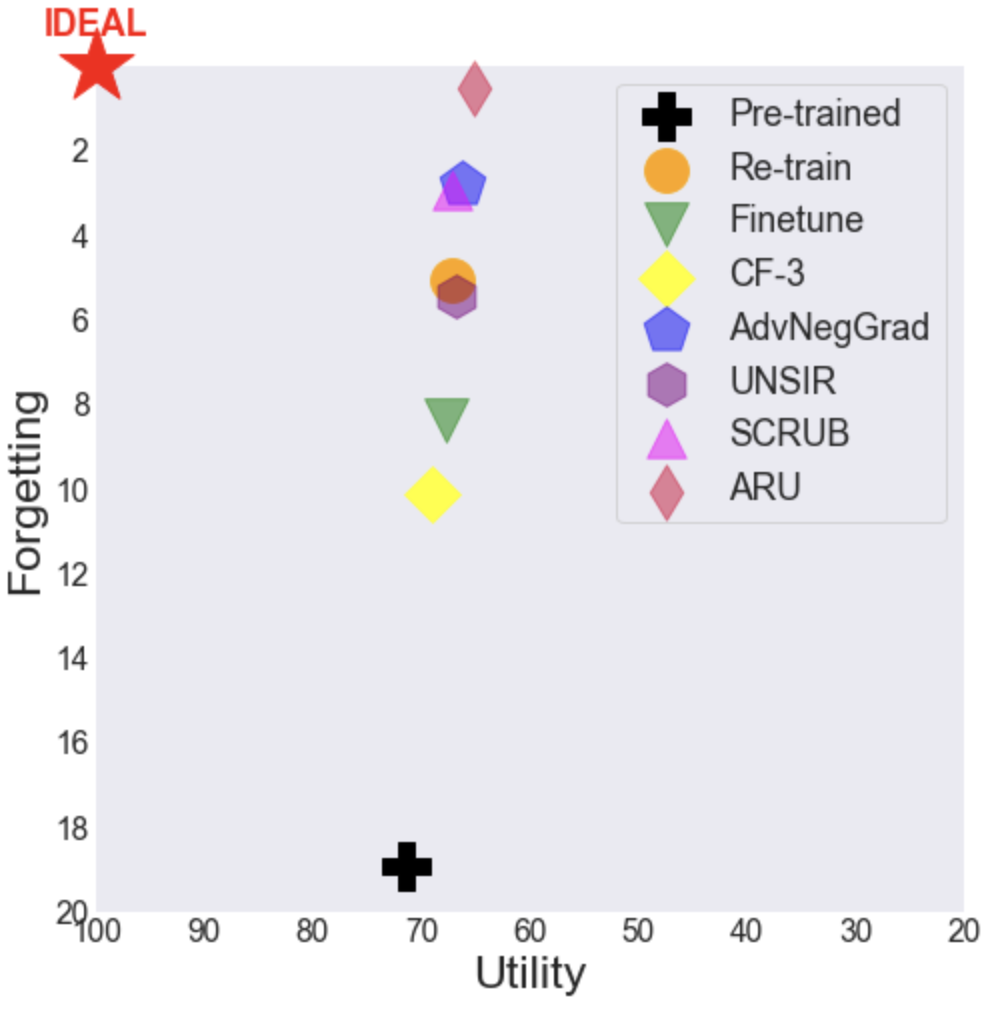}  
  \caption{Trade-off (ViT-L-14, MUFAC)}
\end{subfigure}
\begin{subfigure}{.245\textwidth}
  \centering
  \includegraphics[width=\linewidth]{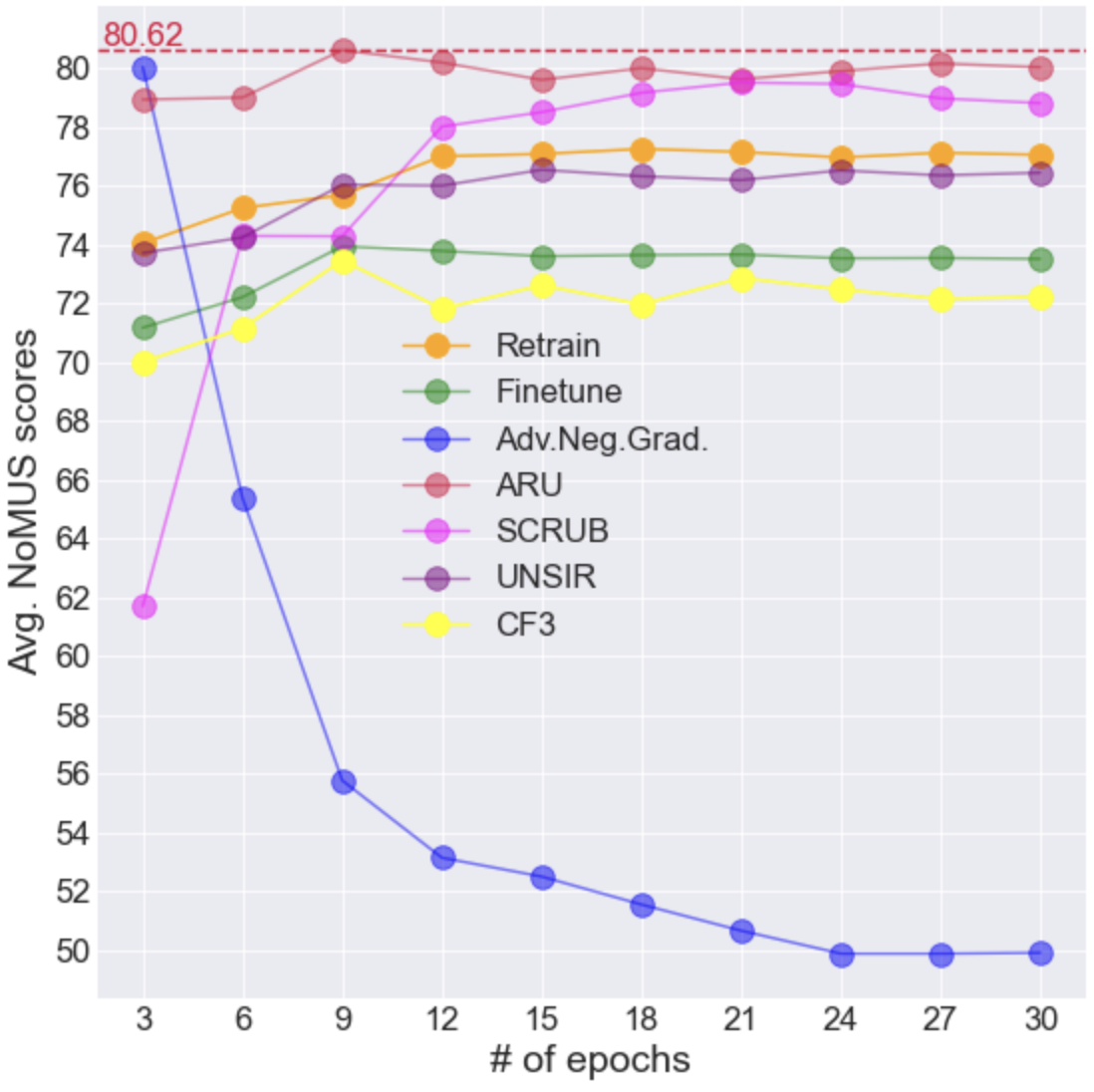}  
  \caption{NoMUS (ViT-L-14, MUFAC)}
\end{subfigure}
\begin{subfigure}{.245\textwidth}
  \centering
  \includegraphics[width=\linewidth]{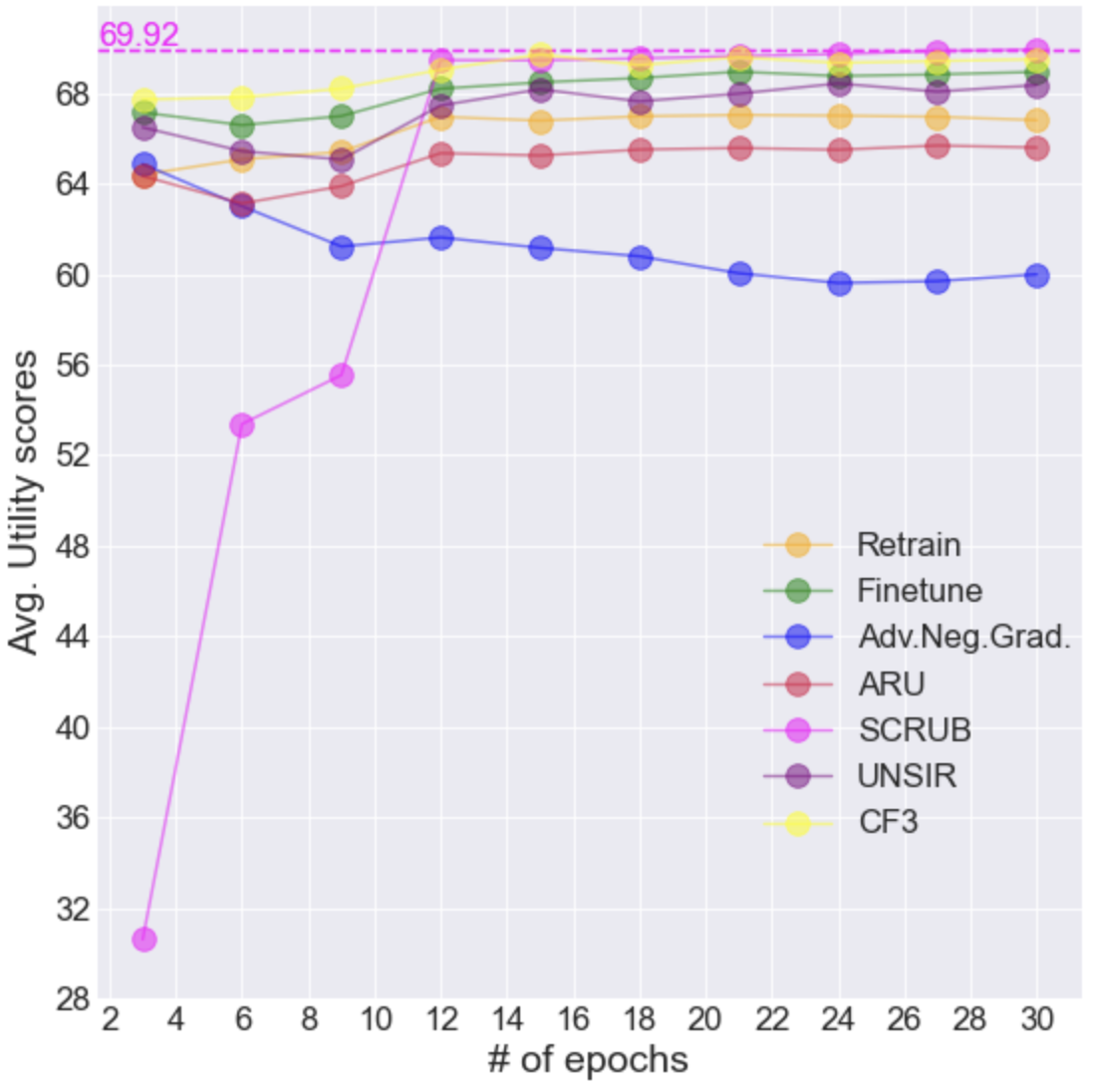}  
  \caption{Utility (ViT-L-14, MUFAC)}
\end{subfigure}
\begin{subfigure}{.245\textwidth}
  \centering
  \includegraphics[width=\linewidth]{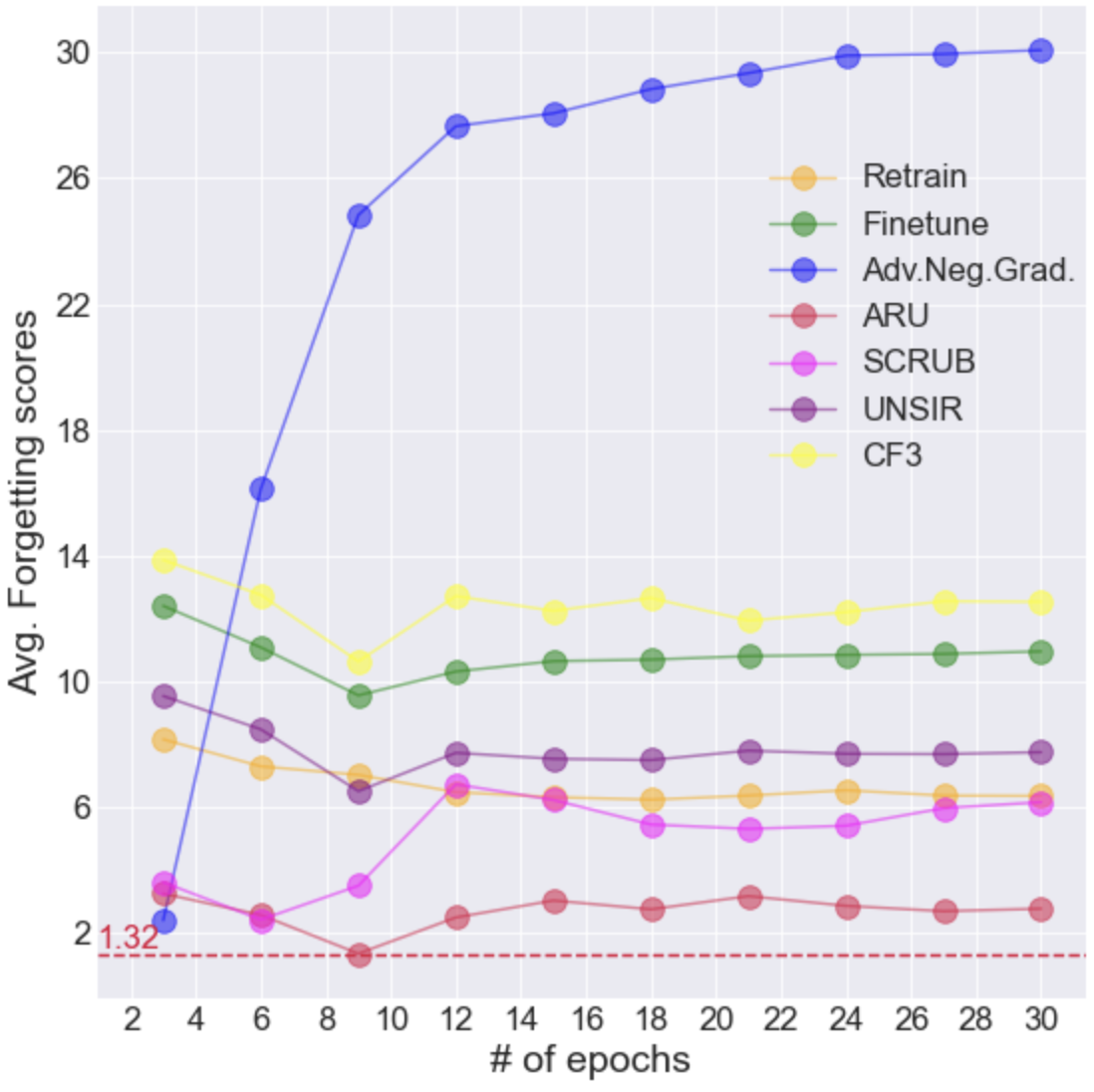}  
  \caption{Forgetting (ViT-L-14, MUFAC)}
\end{subfigure}
\newline
\begin{subfigure}{.245\textwidth}
  \centering
  \includegraphics[width=\textwidth]{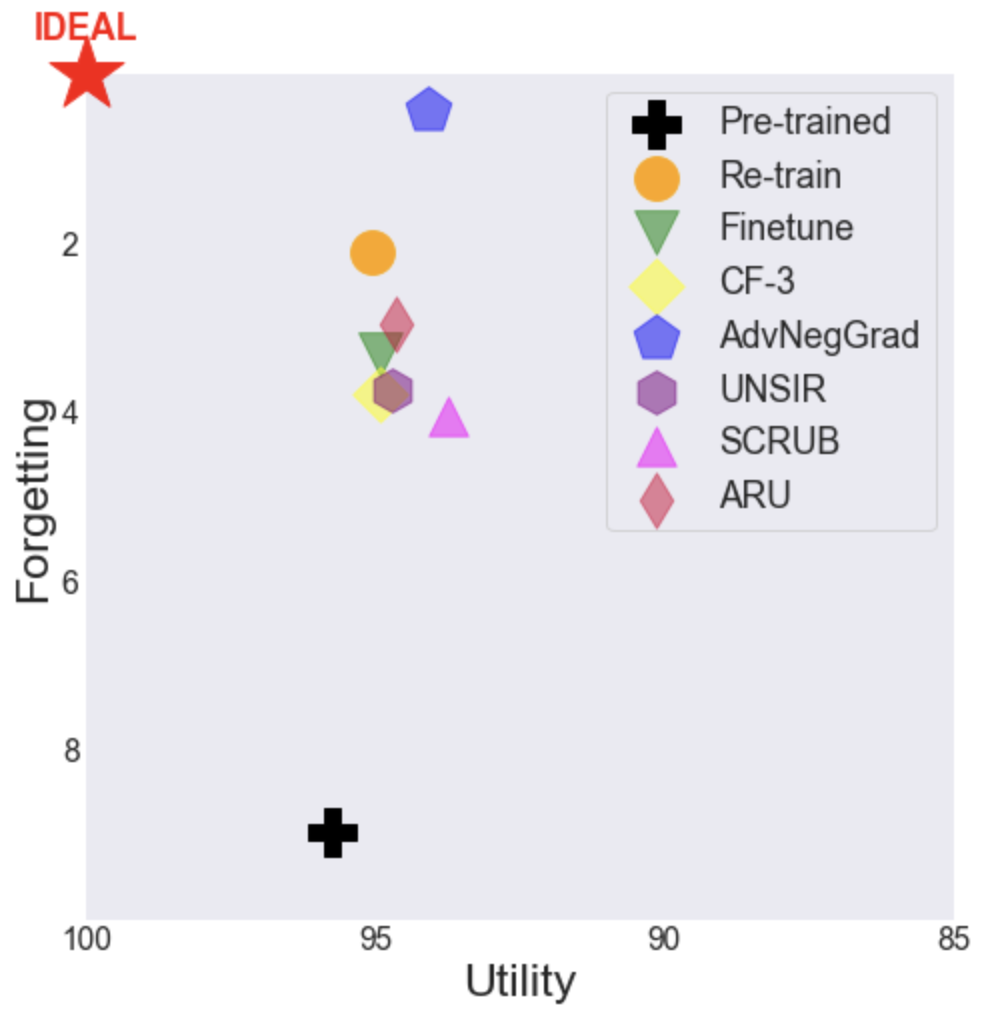}  
  \caption{Trade-off (ViT-B-16, MUCAC)}
\end{subfigure}
\begin{subfigure}{.245\textwidth}
  \centering
  \includegraphics[width=\linewidth]{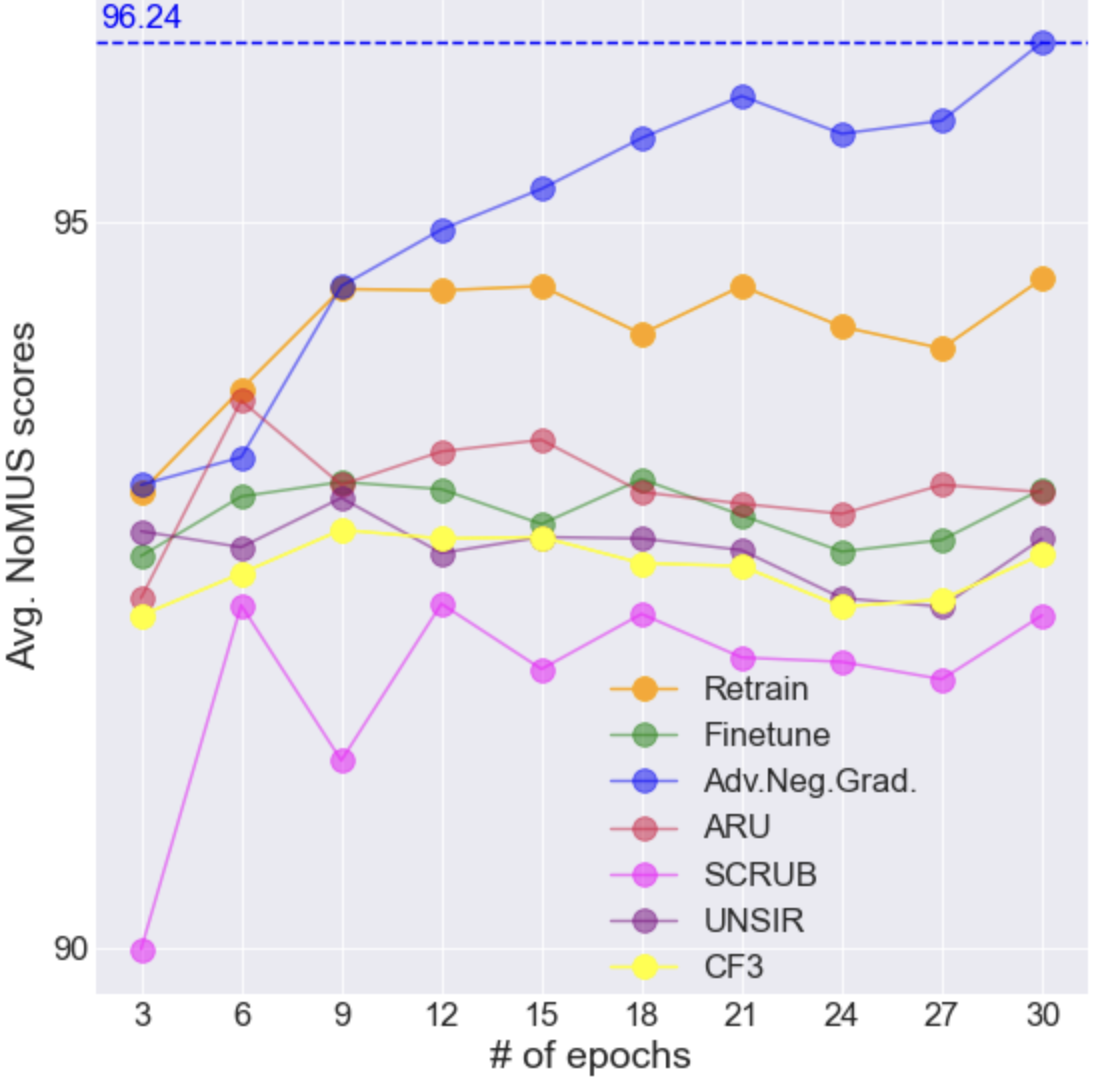}
  \caption{NoMUS (ViT-B-16, MUCAC)}
\end{subfigure}
\begin{subfigure}{.245\textwidth}
  \centering
  \includegraphics[width=\linewidth]{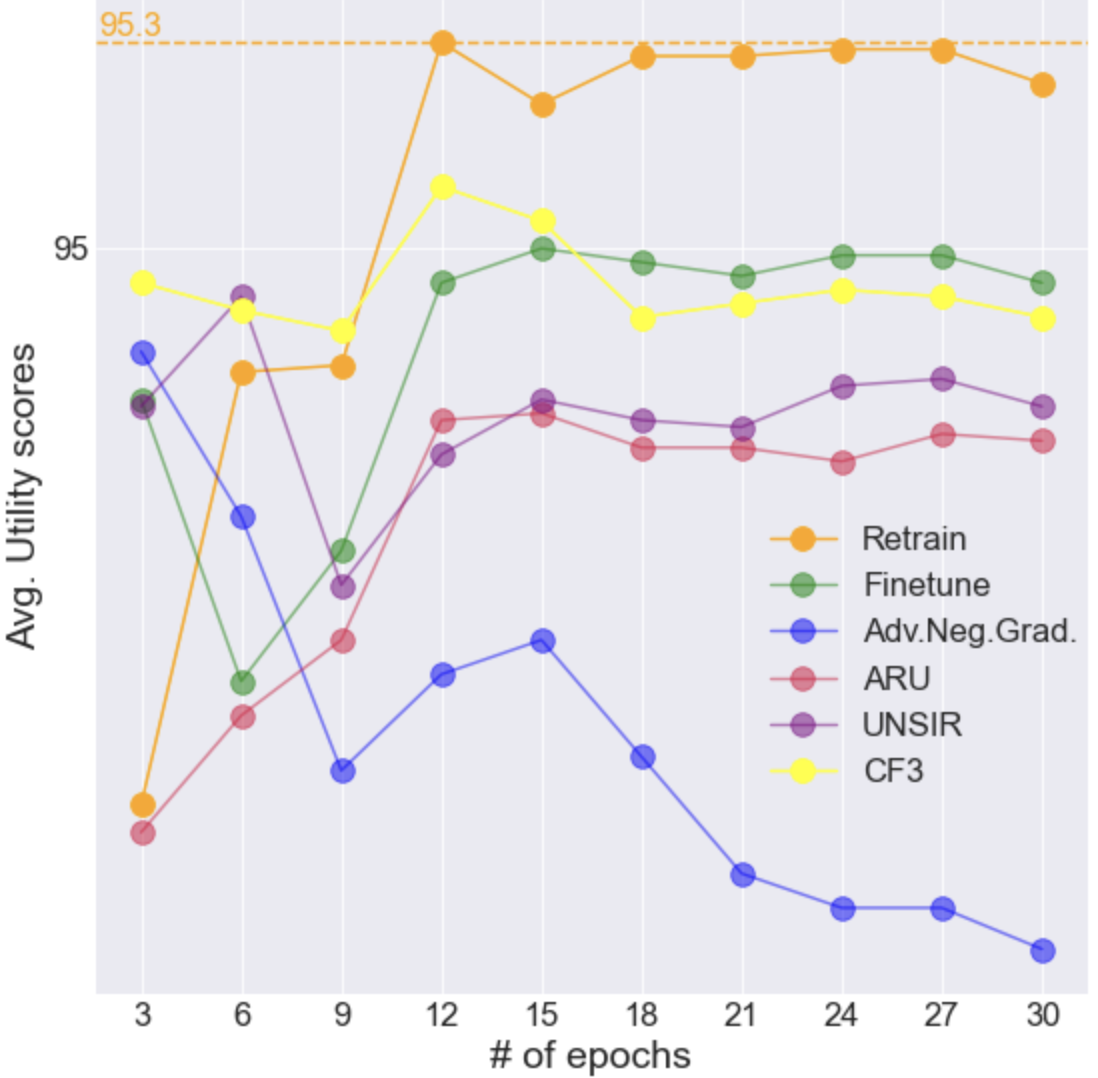}  
  \caption{Utility (ViT-B-16, MUCAC)}
\end{subfigure}
\begin{subfigure}{.245\textwidth}
  \centering
  \includegraphics[width=\linewidth]{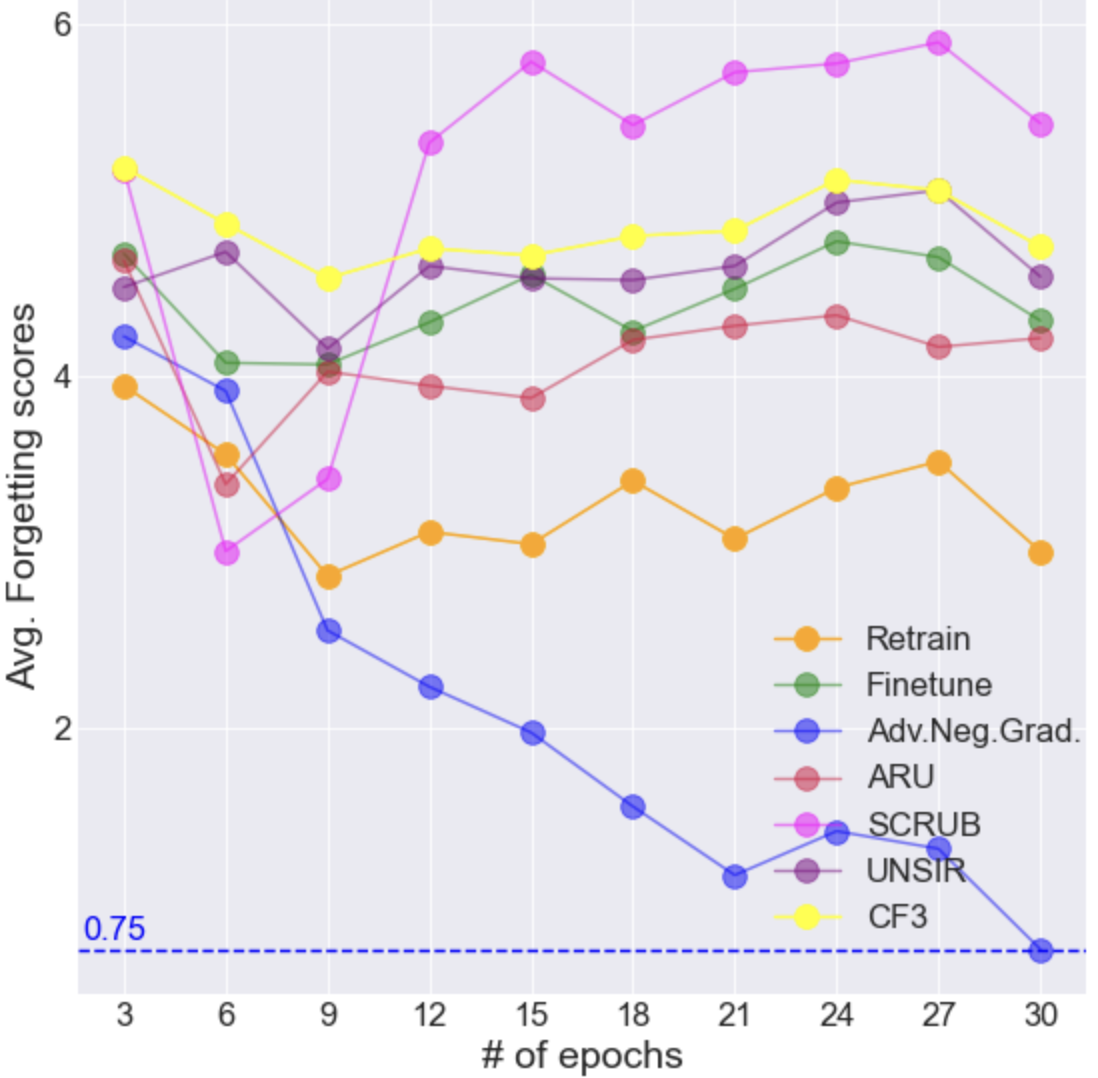} \caption{Forgetting (ViT-B-16, MUCAC)}
\end{subfigure}
\newline
\begin{subfigure}{.245\textwidth}
  \centering
  \includegraphics[width=\textwidth]{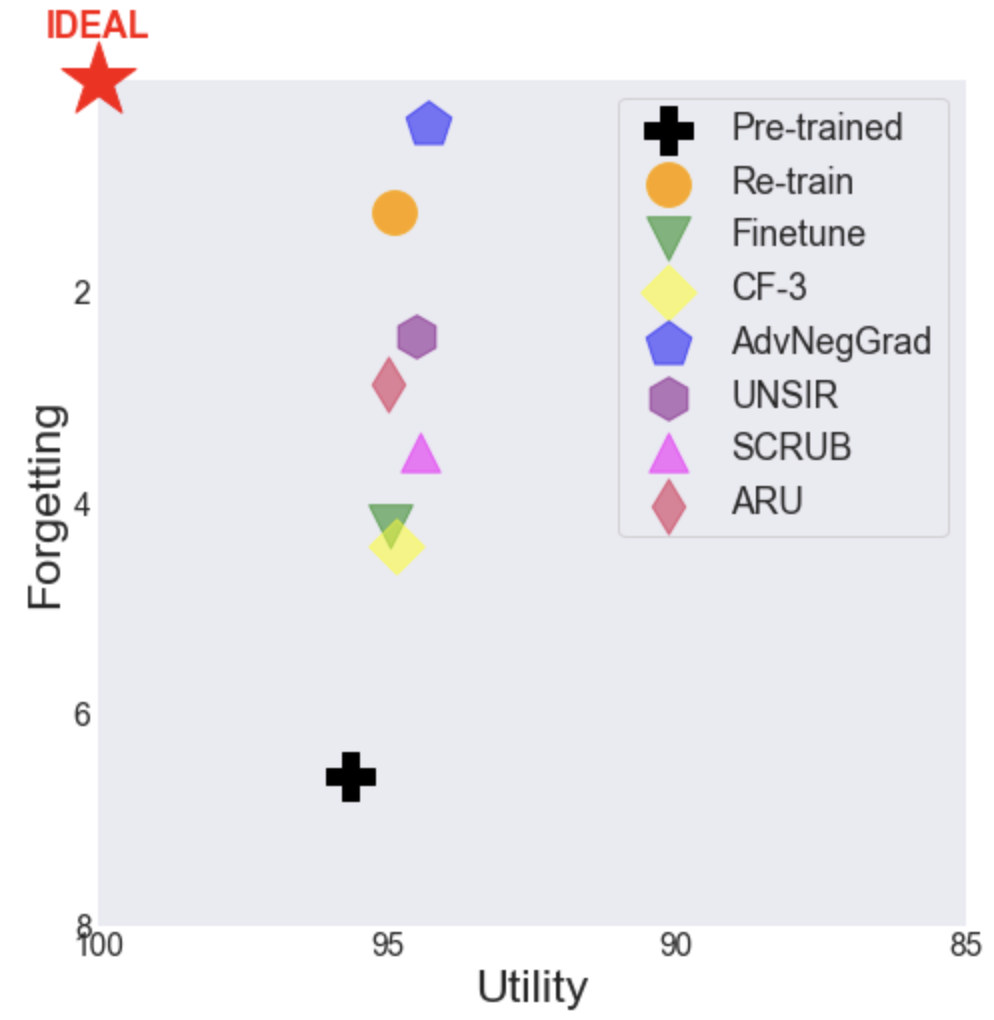}  
  \caption{Trade-off (ViT-L-14, MUCAC)}
\end{subfigure}
\begin{subfigure}{.245\textwidth}
  \centering
  \includegraphics[width=\linewidth]{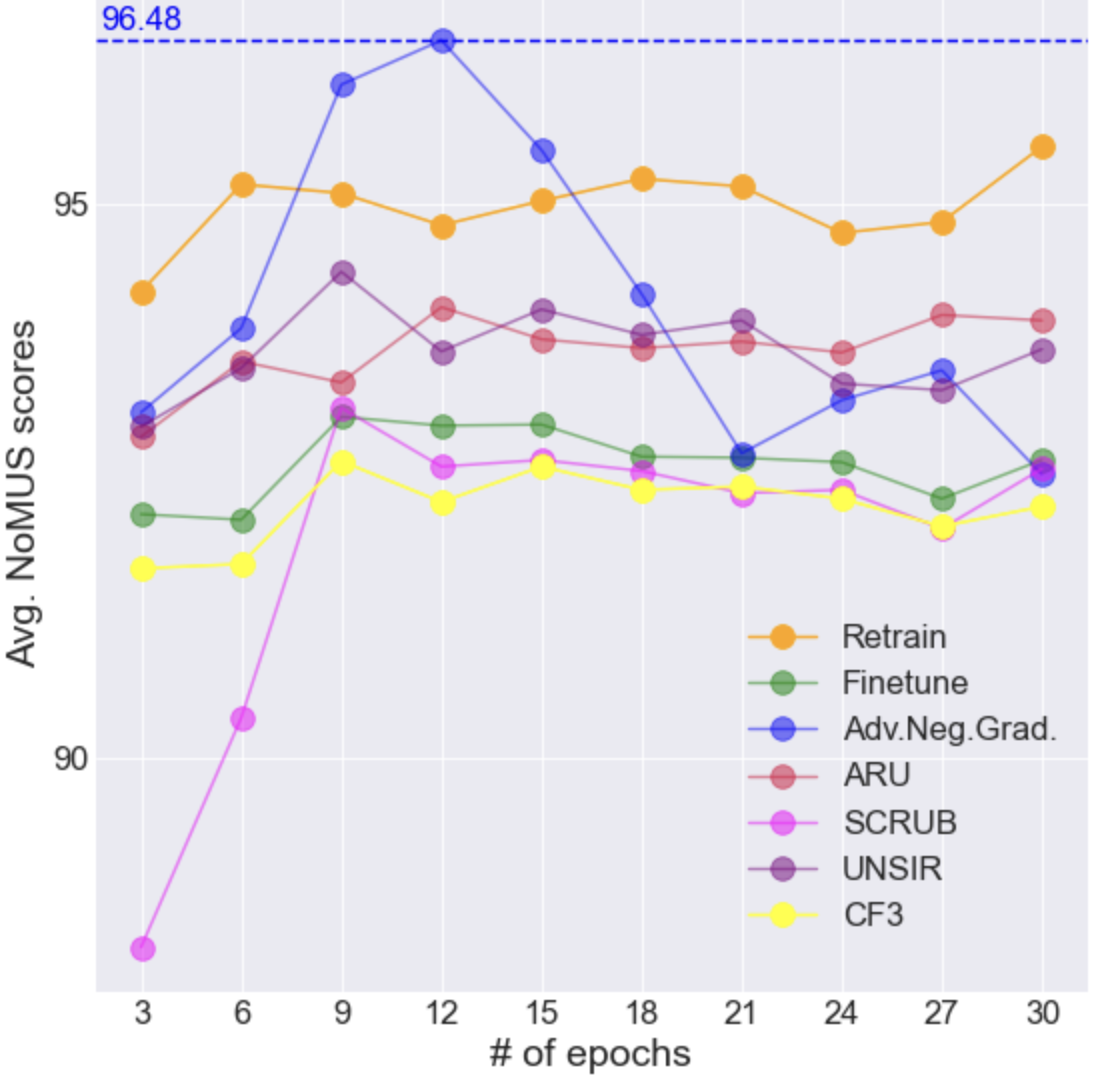}
  \caption{NoMUS (ViT-L-14, MUCAC)}
\end{subfigure}
\begin{subfigure}{.245\textwidth}
  \centering
  \includegraphics[width=\linewidth]{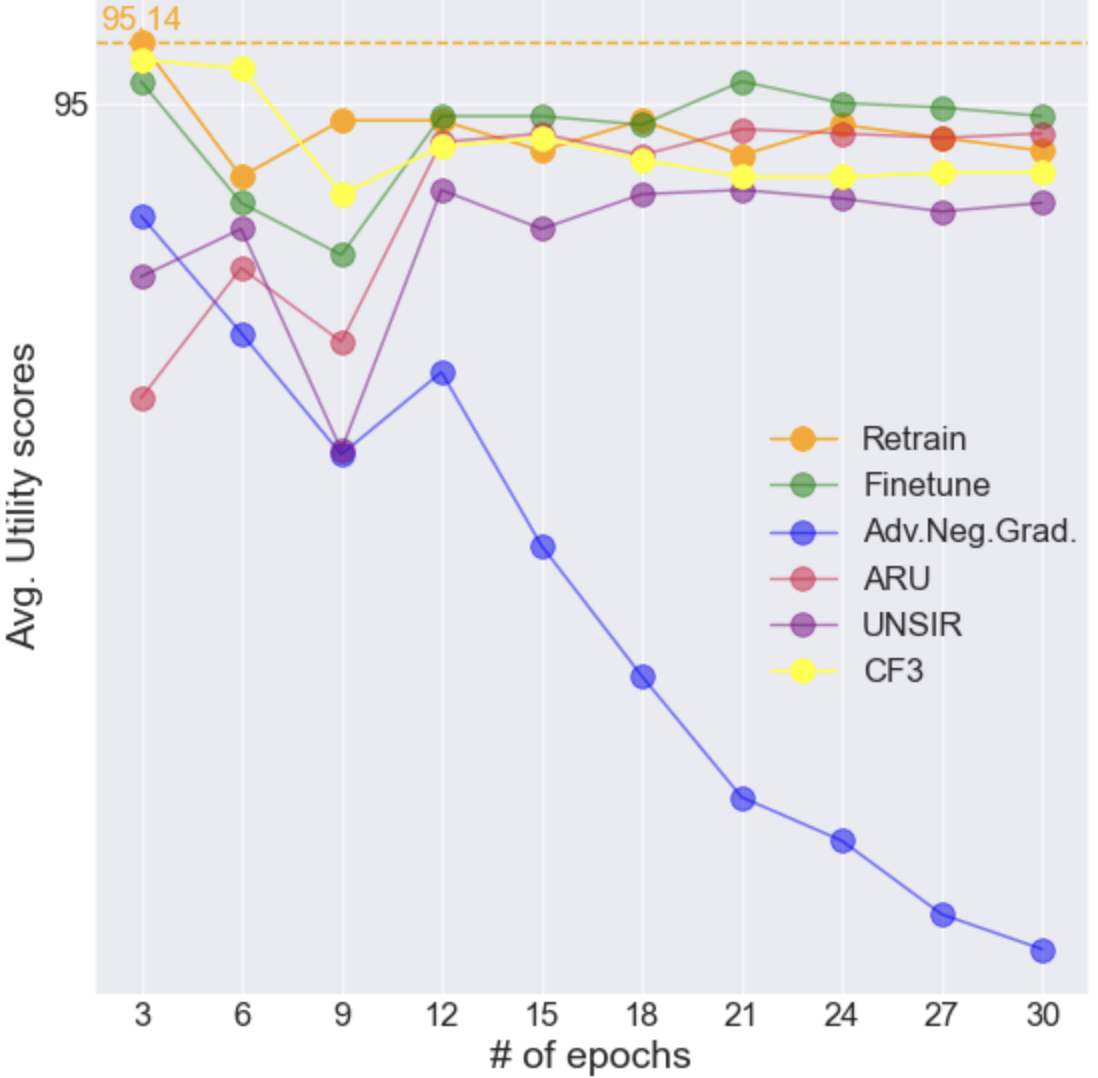}  
  \caption{Utility (ViT-L-14, MUCAC)}
\end{subfigure}
\begin{subfigure}{.245\textwidth}
  \centering
  \includegraphics[width=\linewidth]{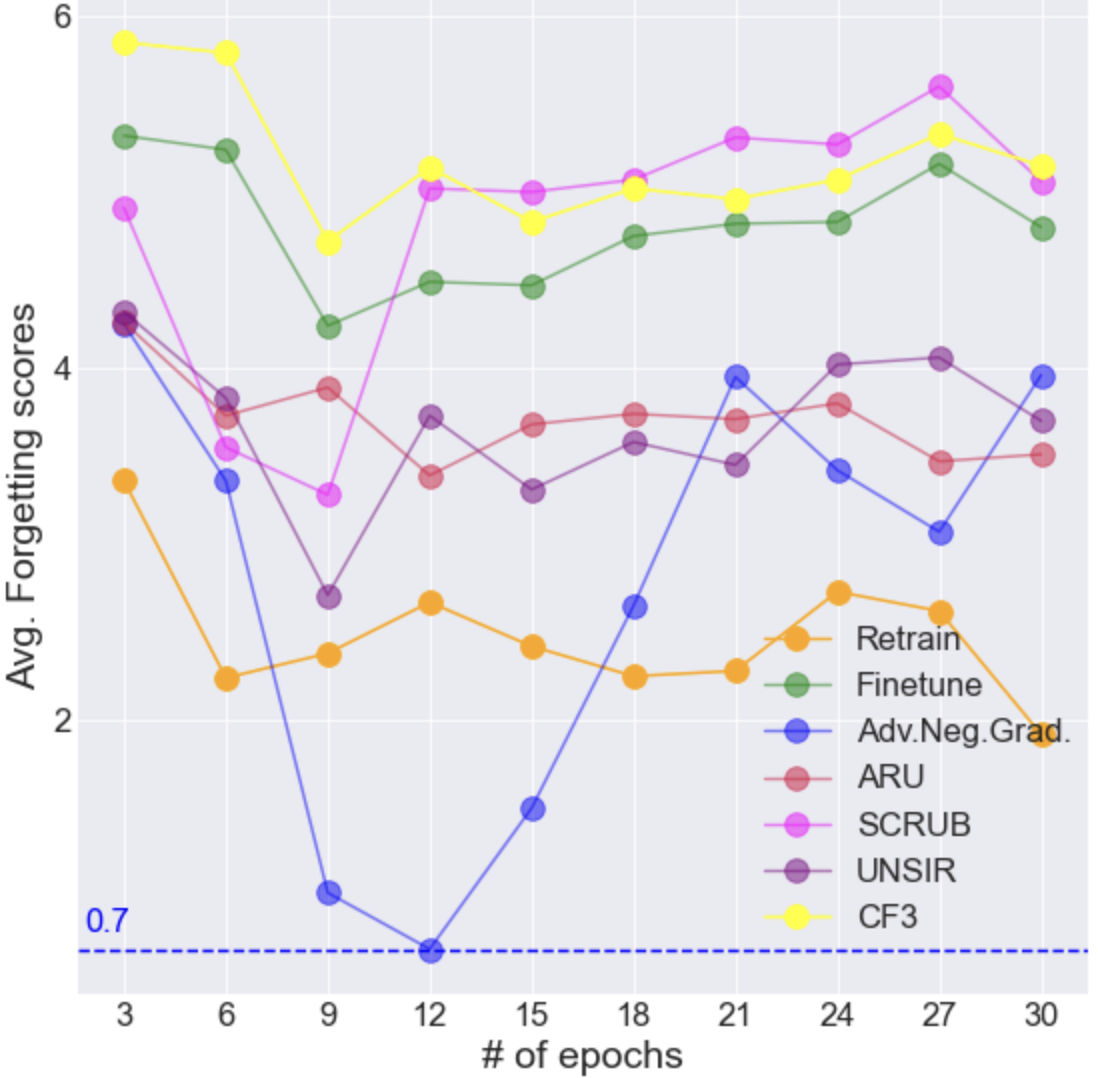} \caption{Forgetting (ViT-L-14, MUCAC)}
\end{subfigure}
\caption{Visualization results of each unlearning algorithms' performance over the epoch. Each mark in the plot represents the average of 5 seed runs. The first column provides a summary of the trade-off between utility and forgetting for each algorithm.}
\label{fig:visual}
\end{figure*}

Table~\ref{tab:overall} presents the results of baseline unlearning techniques applied to ResNet18 and Vision Transformers (ViT-B-16 and ViT-L-14). Based on the results, we observe several findings: (1) Vision Transformers (ViT) show better general performance compared to ResNet18, with ViT-L-14 outperforming ViT-B-16 (e.g., pre-trained models on MUFAC show accuracies of 59.52\% for ResNet18, 66.54\% for ViT-B-16, and 71.35\% for ViT-L-14). This is consistent with the general trend where ViT models are generally superior to ResNets, and the larger ViT-L model outperforms the smaller ViT-B model; (2) All the baseline unlearning methods are effective on ViT models, as evidenced by the increase in the NoMUS score from the pre-trained model; (3) algorithms that are relatively more effective on ResNet18 (e.g., ARU, AdvNegGrad, SCRUB) also show greater effectiveness on ViTs, indicating a general trend between ResNet and ViT.

\subsection{Specific Results}
We provide a concise analysis of each of the unlearning algorithms.

\noindent\textbf{Re-training from scratch}\quad As explained in Section~\ref{020related}, re-training from scratch (denoted by ``Re-train'' in Table~\ref{tab:overall}) serves as the foundational baseline method for machine unlearning. In all four scenarios (\{ViT-B, ViT-L\} x \{MUFAC, MUCAC\}), the re-training method shows improved forgetting performance as expected. However, within our setting, where the forget dataset is relatively large, re-training from scratch generally yields a lower utility score, consequently resulting in a relatively lower overall NoMUS score. Thus, in situations where the forget data is extensive, opting for alternative unlearning algorithms that leverage the pre-trained $\theta_0$ instead of re-training from scratch, could be a wiser choice.

\noindent \textbf{Finetuning on the retain set}\quad 
Fine-tuning on the retain set (indicated as ``Finetune'' in Table~\ref{tab:overall}) proves to be effective in all four settings, particularly in forgetting performance. However, it is relatively less powerful in forgetting compared to re-training. This outcome aligns with expectations since, even though it undergoes fine-tuning on the retain set, the traces of the forget set do not fully vanish.

\noindent \textbf{CF-$k$}\quad
Fine-tuning only the final $k$ layers of the original model on the retain dataset (indicated as ``CF-k'' in Table~\ref{tab:overall}) also demonstrates effectiveness in all four settings. An ablation study, as presented in Table~\ref{tab:aru_ablation}, verifies that larger values of $k$ are more effective. It's worth noting that, in all four cases, CF-k does not outperform simple fine-tuning. We observe that freezing the lower layers is not effective in achieving a satisfactory forgetting performance.

\noindent \textbf{Advanced Negative Gradient}\quad 
Introduced by \cite{choi2023towards}, Advanced Negative Gradient (AdvNegGrad) leverages gradient ascent on the forget set to mitigate overfitting. Consistent with its underlying motivation, AdvNegGrad shows notable forgetting performance on both datasets, for both ViT-B-16 and ViT-L-14 models. These outcomes underscore the effectiveness of integrating gradient ascent for effective forgetting. One limitation is that the model's performance is highly unstable due to the gradient ascent term becoming unboundedly large as unlearning progresses. Balancing the two loss terms and inducing a stable unlearning would be an interesting future work.

\noindent \textbf{UNSIR}\quad Leveraging noise to counteract the overfitting of the model to the forget set (denoted as "UNSIR" in Table~\ref{tab:overall}) proves to be effective in all four cases as well. However, when compared to other strong baselines like AdvNegGrad and SCRUB, the overall performance is slightly inferior.

\noindent \textbf{SCRUB}\quad Similar to AdvNegGrad, the SCRUB method also utilizes gradient ascent on the forget set. SCRUB introduces a coefficient as its hyper-parameter, which balances the two losses (gradient descent on $D_R$ and gradient ascent on $D_F$). We perform a grid search on this coefficient hyperparameter, and the outcomes are summarized in Table~\ref{tab:aru_ablation}. The results highlight the significant influence of this hyperparameter on performance, necessitating the need for hyperparameter search when applying SCRUB to new tasks or models.

\noindent \textbf{ARU}\quad The Attack-and-Reset \cite{jung2024attack} approach is a re-initialization-based method designed to effectively identify and reset parameters that are responsible for overfitting to the forget set. ARU demonstrates state-of-the-art performance on both MUFAC and MUCAC when applied to ResNet18. Similarly, the method is also effective on ViTs. ARU has a unique hyperparameter, the pruning ratio (ranging from 0\% to 100\%), where the final performance heavily depends on. We conducted a fundamental ablation study which involves exploring different resetting ratios (i.e., 10\%, 30\%, 50\%, 70\%, 90\%). We used ViT-B-16 as the representative model.

Table~\ref{tab:aru_ablation} summarizes the results, revealing two main findings: (1) ARU's performance is highly dependent on the pruning ratio when applied to Vision Transformer (ViT) models; (2) pruning more than 10\% of the ViT parameters lead to considerable degradation in performance, while the optimal pruning rate being 50\% for ResNet18. This disparity suggests that ViT-based models might employ parameters more efficiently, leading to lower redundancy.

\begin{table}[h]
  \centering
  \begin{adjustbox}{width=0.45\textwidth,keepaspectratio}
  \begin{tabular}{l|ccc}
    \toprule
    &  & MUFAC & \\
    \midrule
    Model (ViT-B-16)&  Utility (\%, $\uparrow$) & Forget (\%, $\downarrow$) & NoMUS (\%, $\uparrow$)\\
    \midrule
    CF-k \\
    \midrule
    \, $\bullet$ k: 3 & 65.21 ($\pm$0.56) & 7.81 ($\pm$0.64) & 74.79 ($\pm$0.72)\\
    \, $\bullet$ k: 6 & 64.91 ($\pm$1.16) & 4.36 ($\pm$1.07) & 78.09 ($\pm$0.86)\\
    \, $\bullet$ k: 9 & 64.98 ($\pm$0.89) & 2.37 ($\pm$1.27) & 80.12 ($\pm$0.99)\\
    \midrule
    ARU \\
    \midrule
    \, $\bullet$ pruning ratio: 10\% & 62.44 ($\pm$0.62) & 0.96 ($\pm$0.94) & 80.26 ($\pm$0.92)\\
    \, $\bullet$ pruning ratio: 30\%& 53.03 ($\pm$1.56) & 5.95 ($\pm$0.81) & 70.57 ($\pm$0.98)\\
    \, $\bullet$ pruning ratio: 50\%& 37.57 ($\pm$1.33) & 3.76 ($\pm$0.93) & 65.02 ($\pm$0.66)\\
    \, $\bullet$ pruning ratio: 70\%& 31.06 ($\pm$0.95) & 3.91 ($\pm$0.60) & 61.62 ($\pm$0.43)\\
    \, $\bullet$ pruning ratio: 90\%& 30.10 ($\pm$1.05) & 5.07 ($\pm$0.81) & 59.99 ($\pm$1.08)\\
    \midrule
    SCRUB \\
    \midrule
    \, $\bullet$ coefficient: 1.0 & 33.76 ($\pm$5.89) & 4.73 ($\pm$1.34) & 62.15 ($\pm$2.11)\\
    \, $\bullet$ coefficient: 0.1 & 65.93 ($\pm$0.84) & 1.45 ($\pm$0.76) & 81.52 ($\pm$0.42) \\
    \, $\bullet$ coefficient: 0.01 & 63.22 ($\pm$2.73) & 1.71 ($\pm$1.88) & 79.90 ($\pm$1.12)\\
    \bottomrule
  \end{tabular}
  \end{adjustbox}
  \caption{Ablation studies on the effect of different hyperparameter values on ARU and CF-k.}
  \label{tab:aru_ablation}
\end{table}

\subsection{Visualization}
We present visualization results in Figure~\ref{fig:visual} to provide insights into how each algorithm facilitates the unlearning process. We observe several interesting findings: (1) Most algorithms converge to their specific performance after approximately 15 epochs, indicating that running unlearning for 30 epochs could be redundant; (2) AdvNegGrad, which is overall the most effective unlearning algorithm, generally takes more time to reach the peak compared to other methods; (3) Methods that utilize gradient ascent (i.e., SCRUB and AdvNegGrad) become relatively unstable, which we speculate is due to the gradient ascent term easily becoming the dominant loss term.

%% file: 050Conclusion.tex
This paper presents a baseline study on recent machine unlearning approaches applied to Vision Transformer (ViT) models using the recently proposed machine unlearning datasets \cite{choi2023towards}. As ViT models are becoming more and more dominant these days, a comprehensive baseline study of recent unlearning methods on ViT models becomes crucial for future research in this field. We hope our work could provide insights and contribute to an active future research in the field.